\documentclass[runningheads,a4paper]{llncs}

\usepackage{amssymb}
\usepackage{graphicx}
\usepackage{amssymb}
\usepackage{bbding}
\usepackage{subfigure}

\begin{document}

\mainmatter  

\title{RAUNet: Residual Attention U-Net for Semantic Segmentation of Cataract Surgical Instruments}

\titlerunning{Lecture Notes in Computer Science: Authors' Instructions}

%
%
\author{Zhen-Liang Ni\inst{1,2}%
\and Gui-Bin Bian\inst{1,2 (}\Envelope\inst{)}
\and Xiao-Hu Zhou\inst{2}
\and Zeng-Guang Hou\inst{1,2,3}
\and Xiao-Liang Xie\inst{2}
\and Chen Wang\inst{1,2}
\and Yan-Jie Zhou\inst{1,2}
\and Rui-Qi Li\inst{1,2}
\and Zhen Li\inst{1,2}\\}
\authorrunning{}

\institute{
The school of Artificial Intelligence, University of Chinese Academy of Sciences, Beijing 100049, China
\and State Key Laboratory of Management and Control for Complex Systems, Institute of Automation, Chinese Academy of Sciences, Beijing 100190, China
\and CAS Center for Excellence in Brain Science and Intelligence Technology, \\Beijing 100190, China\\
\email{ $\{$nizhenliang2017, guibin.bian$\}$ @ia.ac.cn}
}

%
%
\titlerunning{RAUNet for Semantic Segmentation of Cataract Surgical Instruments}
\authorrunning{Zhen-Liang Ni et al.}
\maketitle

\begin{abstract}
Semantic segmentation of surgical instruments plays a crucial role in robot-assisted surgery. However, accurate segmentation of cataract surgical instruments is still a challenge due to specular reflection and class imbalance issues. In this paper, an attention-guided network is proposed to segment the cataract surgical instrument. A new attention module is designed to learn discriminative features and address the specular reflection issue. It captures global context and encodes semantic dependencies to emphasize key semantic features, boosting the feature representation. This attention module has very few parameters, which helps to save memory. Thus, it can be flexibly plugged into other networks. Besides, a hybrid loss is introduced to train our network for addressing the class imbalance issue, which merges cross entropy and logarithms of Dice loss. A new dataset named Cata7 is constructed to evaluate our network. To the best of our knowledge, this is the first cataract surgical instrument dataset for semantic segmentation. Based on this dataset, RAUNet achieves state-of-the-art performance 97.71$\%$ mean Dice and 95.62$\%$ mean IOU.
\end{abstract}

\section{Introduction}
In recent years, semantic segmentation of surgical instruments has gained increasing popularity due to its promising applications in robot-assisted surgery. One of the crucial applications is the localization and pose estimation of surgical instruments, which contributes to surgical robot control. Potential applications of segmenting surgical instruments include objective surgical skills assessment, surgical workflow optimization, report generation, etc.~\cite{TMI} These applications can reduce the workload of doctors and improve the safety of surgery.

Cataract surgery is the most common ophthalmic surgery in the world. It is performed approximately 19 million times a year~\cite{cata_context}. Cataract surgery is highly demanding for doctors. Computer-assisted surgery can significantly reduce the probability of accidental operation. However, most of the research related to surgical instrument segmentation focuses on endoscopic surgeries. There are few studies on cataract surgeries. To the best of our knowledge, this is the first study to segment and classify cataract surgical instruments.

Recently, a serious of methods have been proposed to segment surgical instruments. Luis \emph{et al}.~\cite{Luis} presented a network based on Fully Convolutional Networks(FCN) and optic flow to solve problems such as occlusion and deformation of surgical instruments. RASNet~\cite{rasnet} adopted an attention module to emphasize the targets region and improve the feature representation. Iro \emph{et al}.~\cite{CSL} proposed a novel U-shape network to provide segmentation and pose estimation of instruments simultaneously. A method combining both recurrent network and the convolutional network was employed by Mohamed \emph{et al}.~\cite{rcnn} to improve the segmentation accuracy. From mentioned above, it can be seen that the convolutional neural network has achieved excellent performance in segmentation of surgical instruments. However, the methods mentioned above are all based on endoscopic surgery. Semantic segmentation of surgical instruments for cataract surgery is quite different from that of endoscopic surgery.

Many challenges need to be faced for the semantic segmentation of cataract surgical instruments. Different from endoscopic surgery, cataract surgery requires strong lighting conditions, leading to serious specular reflection. Specular reflection changes the visual characteristics of surgical instruments. Also, cataract surgery instruments are small for micromanipulation. Hence it is very common that surgical instruments only occupy a small region of the image. The number of background pixels is much larger than that of foreground pixels, which cause serious class imbalance issue. As a result, the surgical instrument is more likely to be misidentified as a background. Occlusion caused by eye tissues and the limited view of the camera is also important issues, causing a part of the surgical instrument to be invisible. These issues make it difficult to identify and segment the surgical instrument.

To address these issues, a novel network, Residual Attention U-Net(RAUNet), is proposed. It introduces an attention module to improve feature representation. The contributions of this work are as follows.
\begin{enumerate}
\item An innovative attention module called augmented attention module (AAM) is designed to efficiently fuse multi-level features and improve feature representation, contributing to addressing the specular reflection issue. Also, it has very few parameters, which helps to save memory.
\item A hybrid loss is introduced to solve the class imbalance issue. It merges cross entropy and logarithm of Dice loss to take advantage of both their merit.
\item To evaluate the proposed network, we construct a cataract surgery instrument dataset named Cata7. As far as we know, this is the first cataract surgery instrument dataset that can be used for semantic segmentation.
\end{enumerate}

\begin{figure}[btp]
\centering
\includegraphics[width=\textwidth]{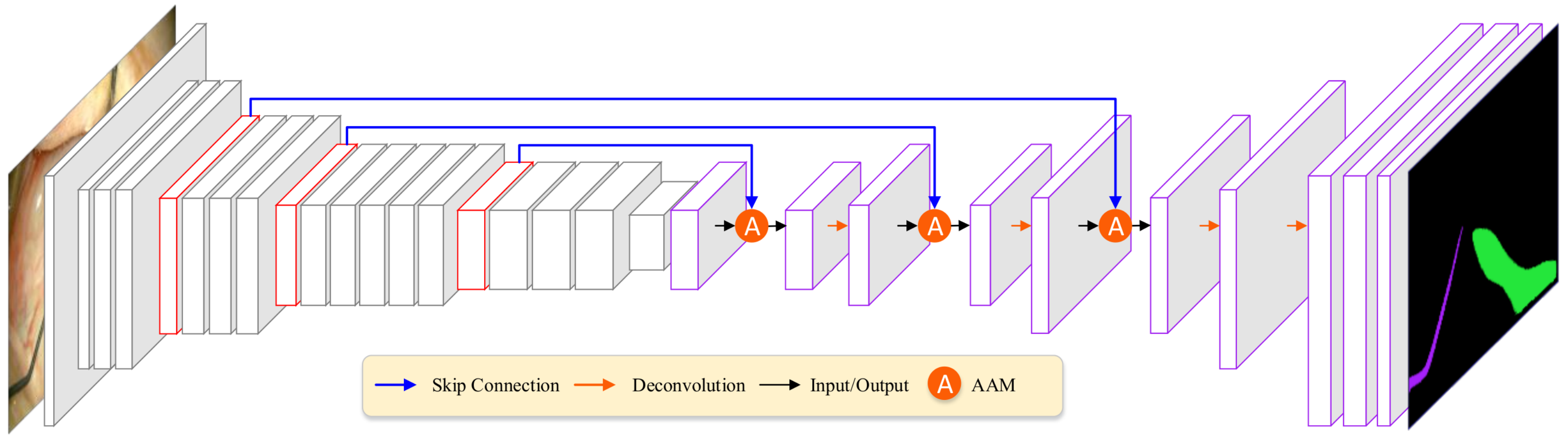}
\caption{The architecture of Residual Attention U-Net. ResNet34 pre-trained on the ImageNet is used as the encoder to capture deep semantic features. The decoder consists of augmented attention module and transposed convolution.}
\label{raunet}
\end{figure}

\section{Residual Attention U-Net}
\subsection{Overview}
High-resolution images provide more detailed location information, helping doctors perform accurate operations. Thus, Residual Attention U-Net (RAUNet) adopts an encoder-decoder architecture to get high-resolution masks. The architecture of RAUNet is illustrated in Fig.~\ref{raunet}. ResNet34~\cite{resnet2016cvpr} pre-trained on the ImageNet is used as the encoder to extract semantic features. It helps reduce the model size and improve inference speed. In the decoder, a new attention module augmented attention module(AAM) is designed to fuse multi-level features and capture global context. Furthermore, transposed convolution is used to carry out upsampling for acquiring refined edges.

\subsection{Augmented Attention Module}
The decoder recovers the position details by upsampling. However, upsampling leads to blurring of edge and the loss of location details. Some existing work~\cite{UNet} adopts skip connections to concatenate the low-level features with the high-level features, which contributes to replenishing the position details. But this is a naive method. Due to the lack of semantic information in low-level features, it contains a lot of useless background information. This information may interfere with the segmentation of the target object. To address this problem, the augmented attention module is designed to capture high-level semantic information and emphasize target features.
\begin{figure}[btp]
\centering
\includegraphics[width=0.95\textwidth]{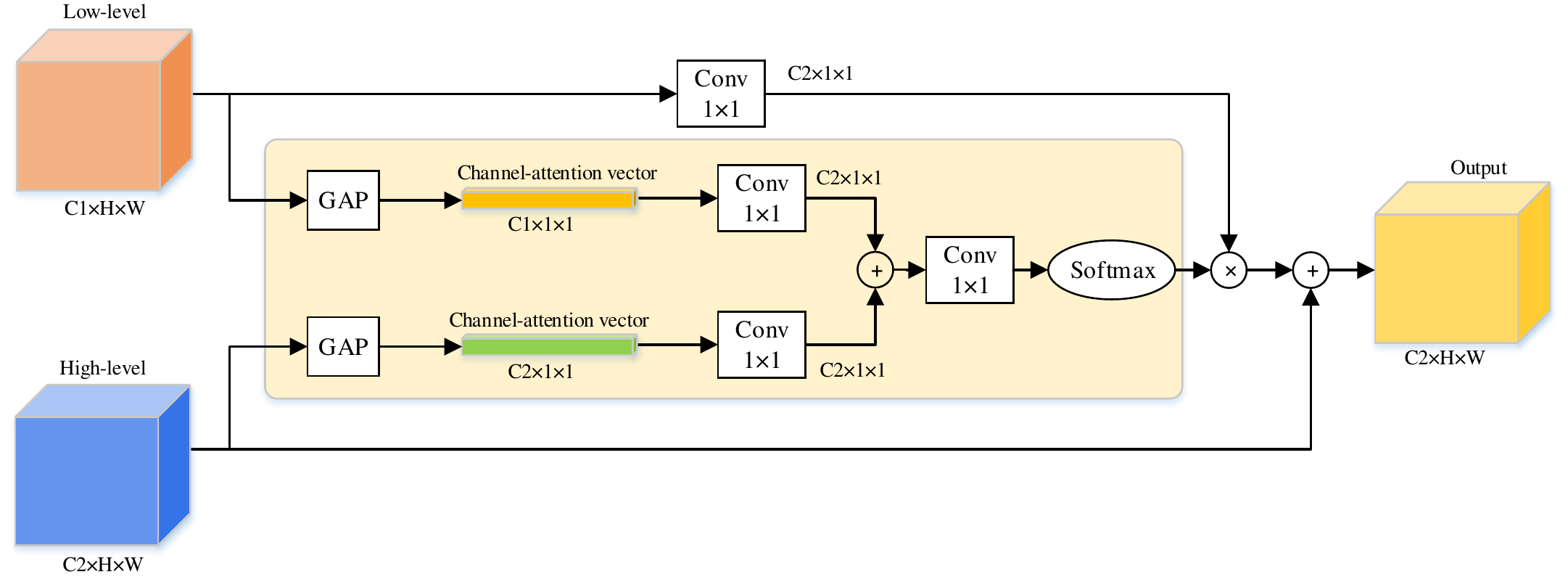}
\caption{The architecture of augmented attention module. $\otimes$ denotes element-wise multiplication. $\oplus$ denotes matrix addition.}
\label{fig1}
\end{figure}

Each channel corresponds to a specific semantic response. Surgical instruments and human tissues are often concerned with different channels. Thus, the augmented attention module model the semantic dependencies to emphasize target channels. It captures the semantic information in high-level feature maps and the global context in the low-level feature maps to encode semantic dependencies. High-level feature maps contain rich semantic information that can be used to guide low-level feature maps to select important location details. Furthermore, the global context of low-level feature maps encodes the semantic relationship between different channels, helping to filter interference information. By using this information efficiently, augmented attention module can emphasize target region and improve the feature representation. Augmented attention module is illustrated in Fig.~\ref{fig1}.

Global average pooling is performed to extract global context and semantic information, which is described in Eq. (3). It squeezes global information into an attentive vector which encodes the semantic dependencies, contributing to emphasizing key features and filter background information. The generation of the attentive vector is described in the following: 
\begin{equation}
{F_a}(x,y) = {\delta _1}\left[ {{W_\alpha }g(x) + {b_\alpha }} \right] + {\delta _1}\left[ {{W_\beta }g(y) + {b_\beta }} \right]
\end{equation}
\begin{equation}
{A_c} = {\delta _2}\left[ {{W_\varphi }{F_a}(x,y) + {b_\varphi }} \right]
\end{equation}
where $x$ and $y$ refer to the high-level and low-level feature maps, respectively. $g$ denotes the global average pooling. ${\delta _1}$ denotes ReLU function and ${\delta _2}$ denotes Softmax function. ${W_\alpha },{W_\beta },{W_\varphi }$ refers to the parameter of the 1$\times$1 convolution. ${b_\alpha },{b_\beta },{b_\varphi }$ refers to the bias.
\begin{equation}
g({x_k}) = \frac{1}{{W \times H}}\sum\limits_{i = 1}^H {\sum\limits_{j = 1}^W {{x_k}(i,j)} }
\end{equation}
where $k = 1,2,...,c$ and ${x} = \left[ {{x_1},{x_2},...,{x_c}} \right]$.

Then 1$\times$1 convolution with batch normalization is performed on the vector to further captures semantic dependencies. The softmax function is adopted as the activation function to normalize the vector. The low-level feature maps are multiplied by the attentive vector to generate an attentive feature map. Finally, the attentive feature map is calibrated by adding with the high-level feature map. Addition can reduce parameters of convolution compared with concatenation, which contributes to reducing the computational cost. Also, since it only uses global average pooling and 1$\times$1 convolution, this module does not add too many parameters. The global average pooling squeezes global information into a vector, which also greatly reduces computational costs.

\subsection{Loss Function}
Semantic segmentation of surgery instruments can be considered as classifying each pixel. Therefore, cross entropy loss can be used for classification of pixels. It is the most commonly used loss function for classification. And it is denoted as \emph{H} in the Eq. (4).
\begin{equation}
H =  - \frac{1}{{w \times h}}\sum\limits_{k = 1}^c {\sum\limits_{i = 1}^w {\sum\limits_{j = 1}^h {{y_{ijk}}} } \log (\frac{{{e^{{{\widehat y}_{ijk}}}}}}{{\sum\limits_{k = 1}^c {{e^{{{\widehat y}_{ijk}}}}} }})}
\end{equation}
where $w$, $h$ represent the width and the height of the predictions. And $c$ is the number of classes. $y_{ijk}$ is the ground truth of a pixel and ${\widehat y}_{ijk}$ is the prediction of a pixel.

It is common that the surgical instrument only occupies a small area of the image, which leads to serious class imbalance issue. However, the performance of cross entropy is greatly affected by this issue. The prediction is more inclined to recognize pixels as background. Therefore the surgical instrument may be partially detected or ignored. The Dice loss defined in Eq. (5) can be used to solve this problem~\cite{soft_dice}. It evaluates the similarity between the prediction and the ground truth, which is not affected by the ratio of foreground pixels to background pixels.

\begin{equation}
D = \frac{{2\sum\limits_i^w {\sum\limits_j^h {{p_{ij}}{g_{ij}}} } }}{{\sum\limits_i^w {\sum\limits_j^h {{p_{ij}}} }  + \sum\limits_i^w {\sum\limits_j^h {{g_{ij}}} } }}
\end{equation}
where $w$, $h$ represent the width and the height of the predictions, $p$ represents the prediction, $g$ represents the ground truth.

To effectively utilize the excellent characteristics of these two losses, we merge the Dice loss with the cross entropy function in the following:
\begin{equation}
L = (1-\alpha)H - \alpha \log (D)
\end{equation}
where $\alpha$ is a weight used to balance cross entropy loss and Dice loss. D is between 0 and 1. $\log (D)$ extends the value range from 0 to negative infinity. When the prediction is greatly different from the ground truth, $D$ is small and $\log (D)$ is close to negative infinity. The loss will increase a lot to penalize this poor prediction. This method can not only use the characteristics of the Dice loss but also improve the sensitivity of loss.

This loss is named Cross Entropy Log Dice(CEL-Dice). It combines the stability of cross entropy and the property that Dice loss is not affected by class imbalance. Therefore, it solves class imbalance better than cross entropy and its stability is better than Dice loss.

\section{Experiments}
\subsection{Datasets}
\begin{figure}[tbp]
\centering
\includegraphics[width=0.94\textwidth]{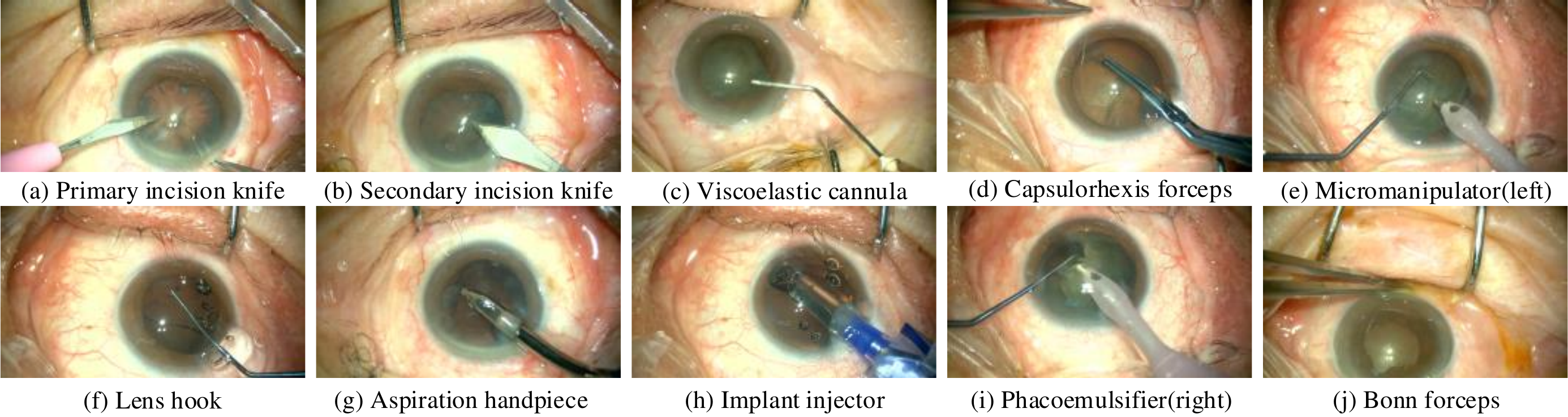}
\caption{Ten cataract surgical instruments used in Cata7. }
\label{cata7}
\end{figure}

\begin{table}[tbp]
\centering
\setlength{\tabcolsep}{2mm}
\caption{Details of Cata7 Dataset.The Cata7 dataset contains a total of 2500 images. It is split into the training set and the test set. Up to two surgical instruments in each image.}
    \begin{tabular}[width=\textwidth]{c|c|c|c|c}
    \hline
    \textbf{No.} & \textbf{Instrument} & \textbf{Number} & \textbf{Train} & \textbf{Test} \\
    \hline
    1     & Primary Incision Knife & 62    & 39    & 23 \\
    2     & Secondary Incision Knife & 226   & 197   & 29 \\
    3     & Viscoelastic Cannula & 535   & 420   & 115 \\
    4     & Capsulorhexis Forceps & 119   & 34    & 85 \\
    5     & Micromanipulator & 507   & 366   & 141 \\
    6     & Lens Hook & 475   & 405   & 70 \\
    7     & Aspiration Handpiece & 515   & 364   & 151 \\
    8     & Implant Injector & 303   & 277   & 26 \\
    9     & Phacoemulsifier Handpiece & 677   & 526   & 151 \\
    10    & Bonn Forceps & 222   & 101   & 121 \\
    \hline
    -     & Total & 3641  & 2729  & 912 \\
    -     & Number of Frames & 2500  & 1800  & 700 \\
    \hline
    \end{tabular}%
  \label{dataset}%
\end{table}%

A new dataset, Cata7, is constructed to evaluate our network, which is the first cataract surgical instrument dataset for semantic segmentation. The dataset consists of seven videos while each video records a complete cataract surgery. All videos are from Beijing Tongren Hospital. Each video is split into a sequence of images, where resolution is 1920$\times$1080 pixels. To reduce redundancy, the videos are downsampled from 30 fps to 1 fps. Also, images without surgical instruments are manually removed. Each image is labeled with precise edges and types of surgical instruments.

This dataset contains 2,500 images, which is divided into training and test sets. The training set consists of five video sequences and test set consists of two video sequences. The number of surgical instruments in each category is illustrated in Table~\ref{dataset}. There are ten surgical instruments used in the surgery, which are shown in Fig.~\ref{cata7}.

\subsection{Training}
ResNet34 pre-trained on the ImageNet is utilized as the encoder. Pre-training can accelerate network convergence and improve network performance~\cite{Ternausnet}. Due to limited computing resources, each image for training is resized to 960$\times$544 pixels. The network is trained by using Adam with batch size 8. The learning rate is dynamically adjusted during training to prevent overfitting. The initial learning rate is $4 \times {10^{ - 5}}$. For every 30 iterations, the learning rate is multiplied by 0.8. As for the $\alpha$ in the CEL-Dice, it is set to 0.2 after several experiments. Dice coefficient and Intersection-Over-Union(IOU) are selected as the evaluation metric.

Data augmentation is performed to prevent overfitting. The augmented samples are generated by random rotation, shifting and flipping. 800 images are obtained by data augmentation, increasing feature diversity to prevent over-fitting effectively. Batch normalization is used for regularization. In the decoder, batch normalization is performed after each convolution.

\subsection{Results}
\subsubsection{Ablation for augmented attention module}
Augmented attention module(AAM) is designed to aggregate multi-level features. It captures global context and semantic dependencies to emphasize key features and suppress background features. To verify its performance, we set up a series of experiments. The results are shown in the Table~\ref{ablation_aam}.
\begin{table}[htbp]
\setlength{\tabcolsep}{2.5mm}
  \centering
  \caption{Performance comparison of attention module(AM).}
    \begin{tabular}{c|c|c|c|c}
    \hline
    \textbf{Method} & \textbf{AM} & \textbf{mDice(\%)} & \textbf{mIOU(\%)}& \textbf{Param}\\
    \hline
    BaseNet & --    & 95.12 & 91.31 & 21.80M\\
    BaseNet & AAM(Ours)   & 97.71 & 95.62 &22.06M\\
    BaseNet & GAU~\cite{PAN} & 96.61 & 93.76& 22.66M\\
    \hline
    \end{tabular}%
  \label{ablation_aam}%
\end{table}%

RAUNet without AAM is used as the base network, which achieves 95.12$\%$ mean Dice and 91.31$\%$ mean IOU. The base network with AAM achieves 97.71$\%$ mean Dice and 95.62$\%$ mean IOU. By applying AAM, mean Dice increases by 2.59$\%$ and mean IOU increases by 4.31$\%$. Furthermore, AAM is compared with GAU~\cite{PAN}. The base network with GAU achieves 96.61$\%$ mean Dice and 93.76$\%$ mean IOU. Compared to the base network with AAM, its mean Dice and mean IOU are reduced by 1.10\% and 1.86\%, respectively. Besides, by applying AAM, parameters only increase by 0.26M, which is 1.19$\%$ of the base network. By applying GAU, parameters increase by 0.60M, which is 2.31 times the amount of parameters increased by AAM. These results show that AAM can not only significantly increase the segmentation accuracy, but also does not add too many parameters.

\begin{figure}[tbp]
\centering
\includegraphics[width=\textwidth]{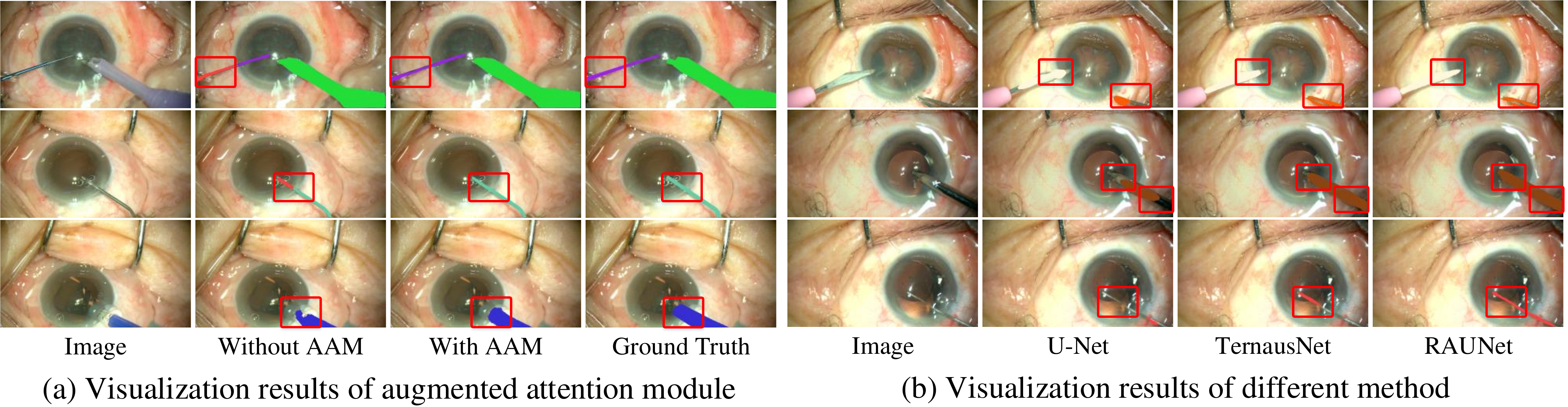}
\caption{Visualization of segmentation results. (a) the visualization results of augmented attention module. (b) the visualization results of various method.}
\label{aam_com}
\end{figure}

To give an intuitive comparison, the segmentation results of the base network and RAUNet are visualized in Fig.~\ref{aam_com}(a). The red line marks the contrasted region. It can be found that there are classification errors in the results of the base network. Besides, surgical instruments are not entirely segmented in the third image. Meanwhile, RAUNet can accurately segment surgical instruments by applying AAM. The masks achieved by RAUNet are the same as the ground truth. This shows that AAM contributes to capturing high-level semantic features and improving feature representation.

\subsubsection{Comparison with state-of-the-art}
To further verify the performance of RAUNet, it is compared with the U-Net~\cite{UNet}, TernausNet~\cite{Ternausnet} and LinkNet~\cite{LinkNet}. As shown in Table~\ref{compare}, RAUNet achieves state-of-the-art performance 97.71$\%$ mean Dice and 95.62$\%$ mean IOU, which outperforms other methods. U-Net~\cite{UNet} achieves 94.99$\%$ mean Dice and 91.11$\%$ mean IOU. TernausNet~\cite{Ternausnet} and LinkNet~\cite{LinkNet} achieve 92.98$\%$ and 92.21$\%$ mean IOU respectively. The performance of these methods is much poor than our RAUNet.
\begin{table}[htbp]
\setlength{\tabcolsep}{3mm}
  \centering
  \caption{Performance comparison of various methods on Cata7.}
    \begin{tabular}{c|c|c|c}
    \hline
    \textbf{Method} & \textbf{mDice(\%)} & \textbf{mIOU(\%)} & \textbf{Param} \\
    \hline
    U-Net~\cite{UNet} & 94.99    &  91.11 & 7.85M \\
    LinkNet~\cite{Ternausnet} &    95.75   &    92.21   & 21.80M \\
    TernausNet~\cite{LinkNet} &   96.24    &   92.98    & 25.36M \\
    \hline
    RAUNet(Ours) & 97.71 & 95.62 &22.06M\\
    \hline
    \end{tabular}%
  \label{compare}%
\end{table}%

\begin{figure}[tbp]
\centering
\subfigure[RAUNet]{
\begin{minipage}{0.35\textwidth}
\centering
\includegraphics[width=\textwidth]{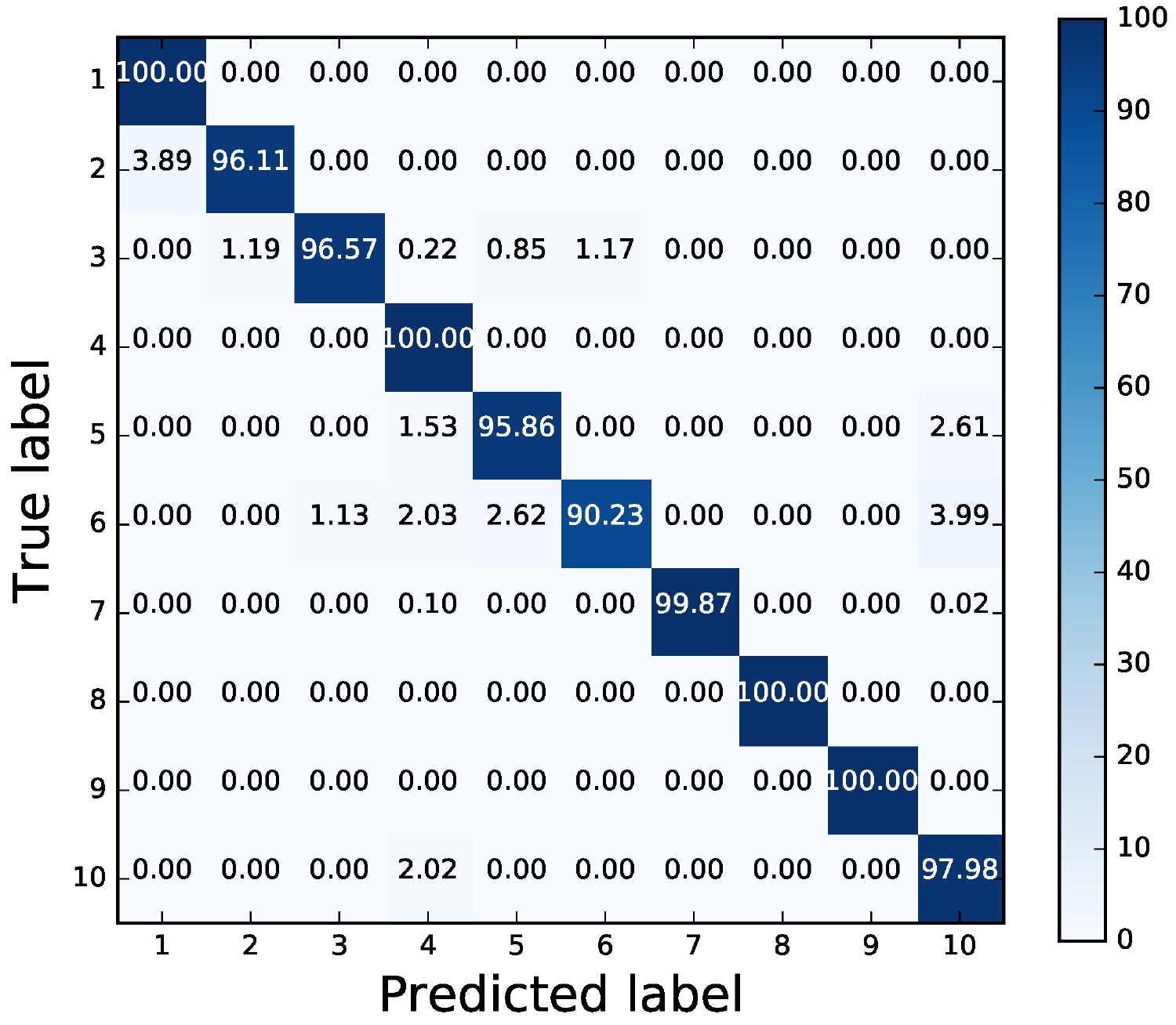}
\end{minipage}
}
\subfigure[U-Net]{
\begin{minipage}{0.35\textwidth}
\centering
\includegraphics[width=\textwidth]{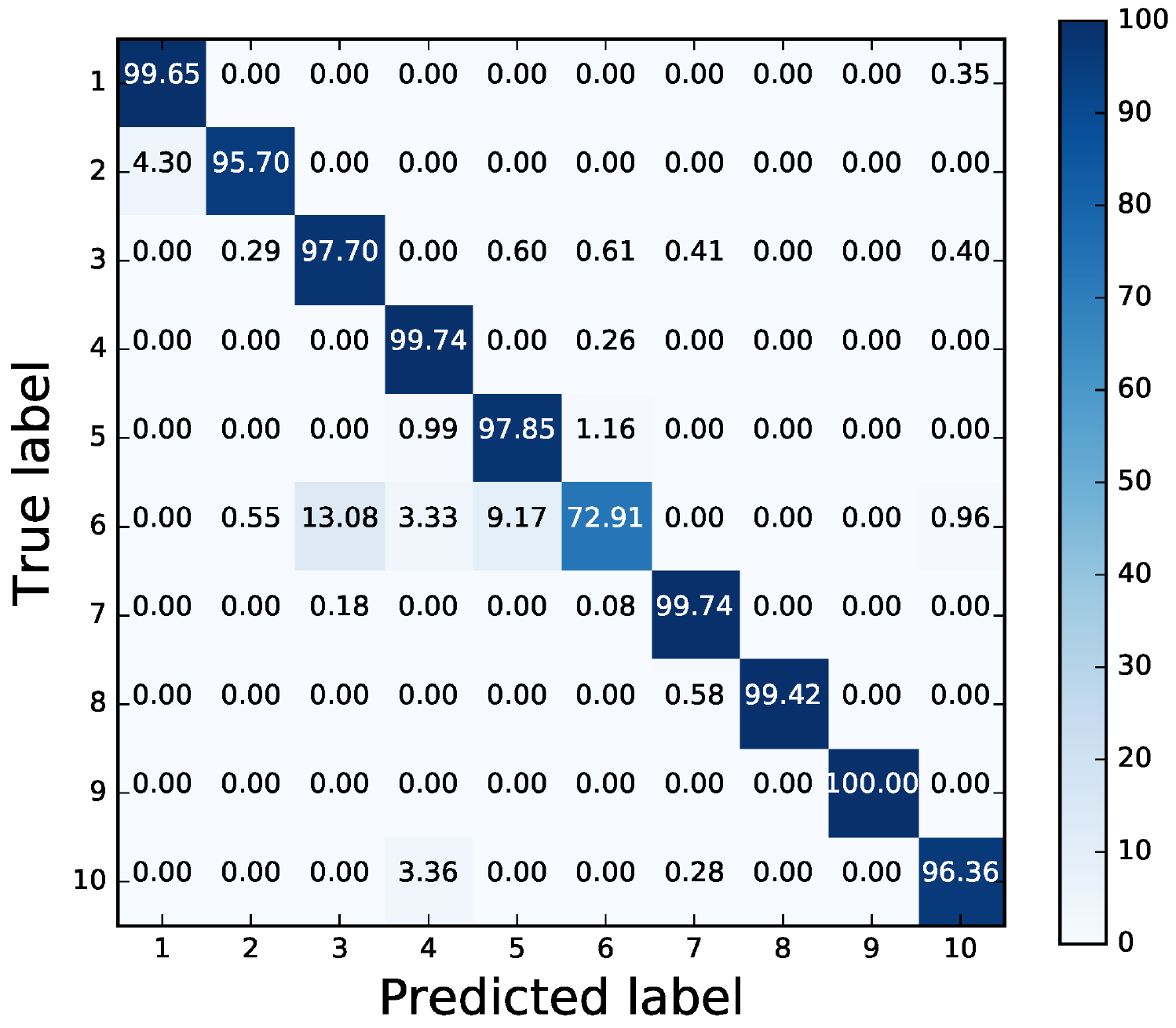}
\end{minipage}
}
\subfigure[TernausNet]{
\begin{minipage}{0.35\textwidth}
\centering
\includegraphics[width=\textwidth]{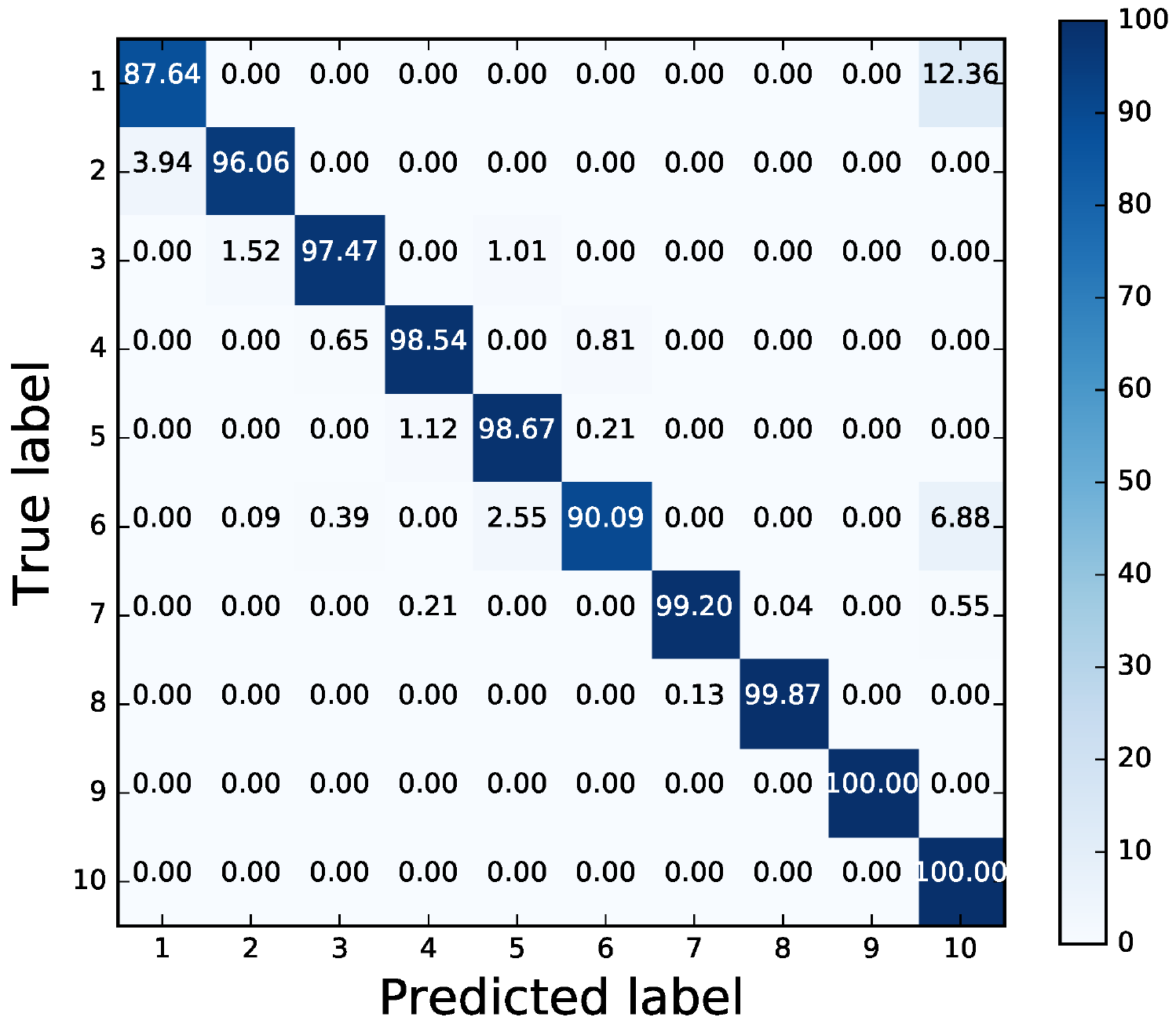}
\end{minipage}
}
\subfigure[LinkNet]{
\begin{minipage}{0.35\textwidth}
\centering
\includegraphics[width=\textwidth]{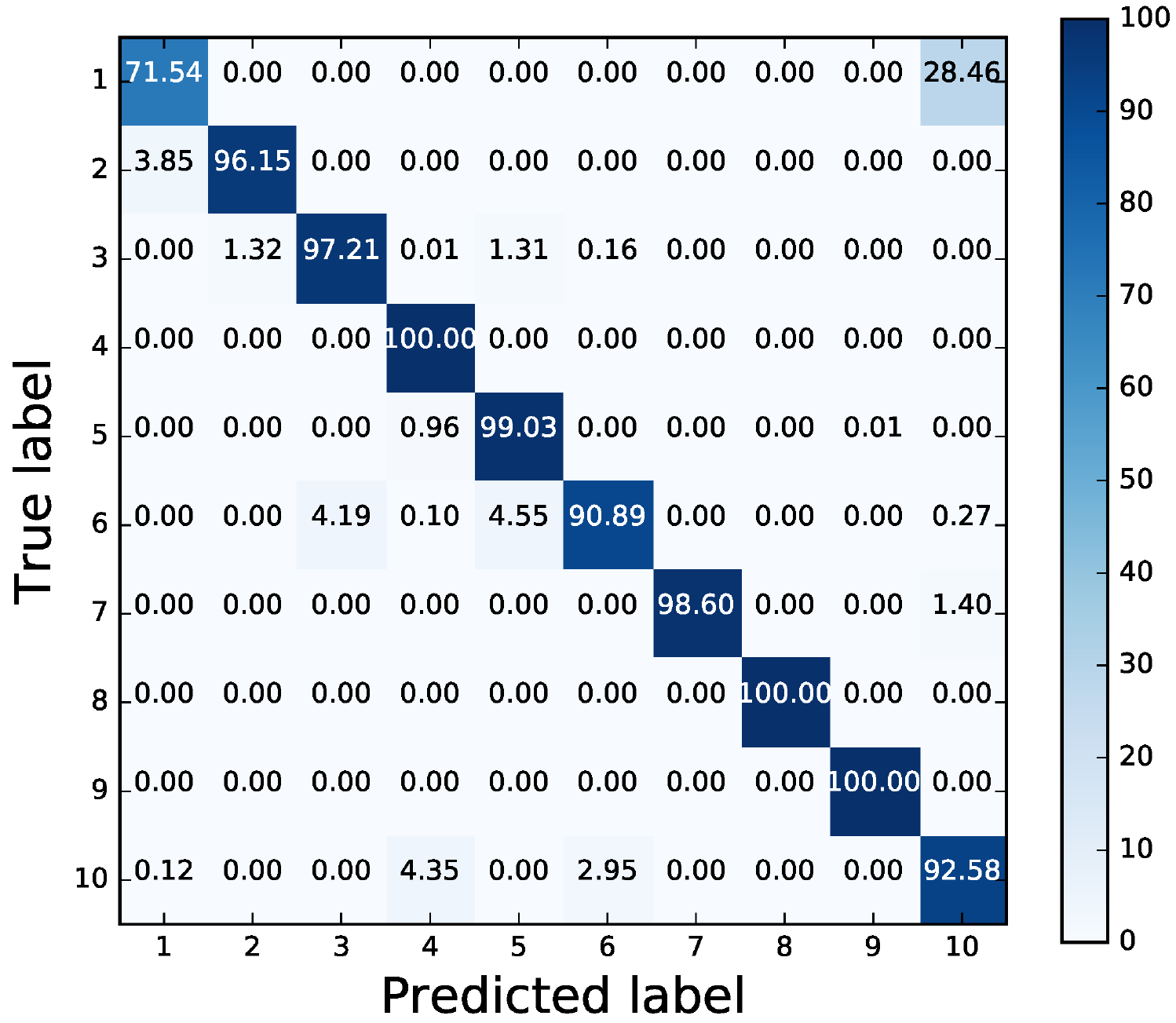}
\end{minipage}
}
\caption{Pixel accuracy of each class. The numbers on the axes represent the categories of surgical instruments.}
\label{confusion}
\end{figure}

Pixel accuracy achieved by various methods is visualized in Fig.~\ref{confusion}. It can be found that the primary incision knife is often misclassified by U-Net, TernausNet, and LinkNet. Since the primary incision knife is used for a short time in surgery, its samples are few, leading to the underfitting of the network. Also, lens hook is often misclassified by U-Net. This result is since the lens hook is very thin and cause severe class imbalance. Furthermore, it is similar to other surgical instruments. U-Net cannot capture high-level semantic information, which causes the misclassification. Despite these difficulties, our method still achieves high pixel accuracy. The pixel accuracy of lens hook and primary incision knife are 90.23$\%$ and 100$\%$ respectively. These results show that RAUNet can capture discriminative semantic features and address the class imbalance issue.

\begin{figure}[htbp]
\centering
\includegraphics[width=\textwidth]{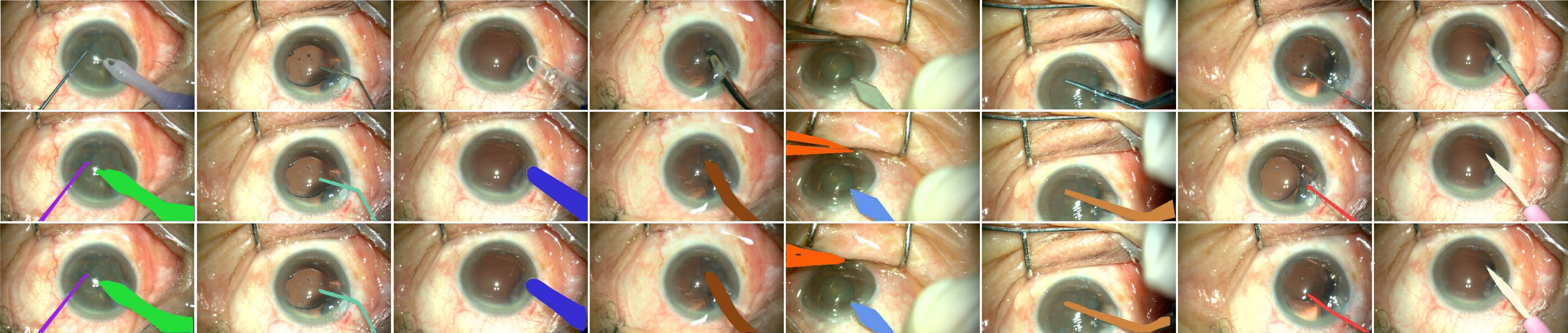}
\caption{Visualization results of ten cataract surgical instruments. From top to low: original image, ground truth and prediction of RAUNet.}
\label{fig4}
\end{figure}

To give an intuitive result, the segmentation results of the above method are visualized in Fig.~\ref{aam_com}(b). The segmentation results of RAUNet are the same as the ground truth, which is significantly better than other methods. Also, more results of RAUNet are shown in Fig.~\ref{fig4}.

\subsubsection{Verify the performance of CEL-Dice}
\begin{figure}[htbp]
\centering
\subfigure[The prediction of mean Dice]{
\begin{minipage}{0.46\textwidth}
\centering
\includegraphics[width=\textwidth]{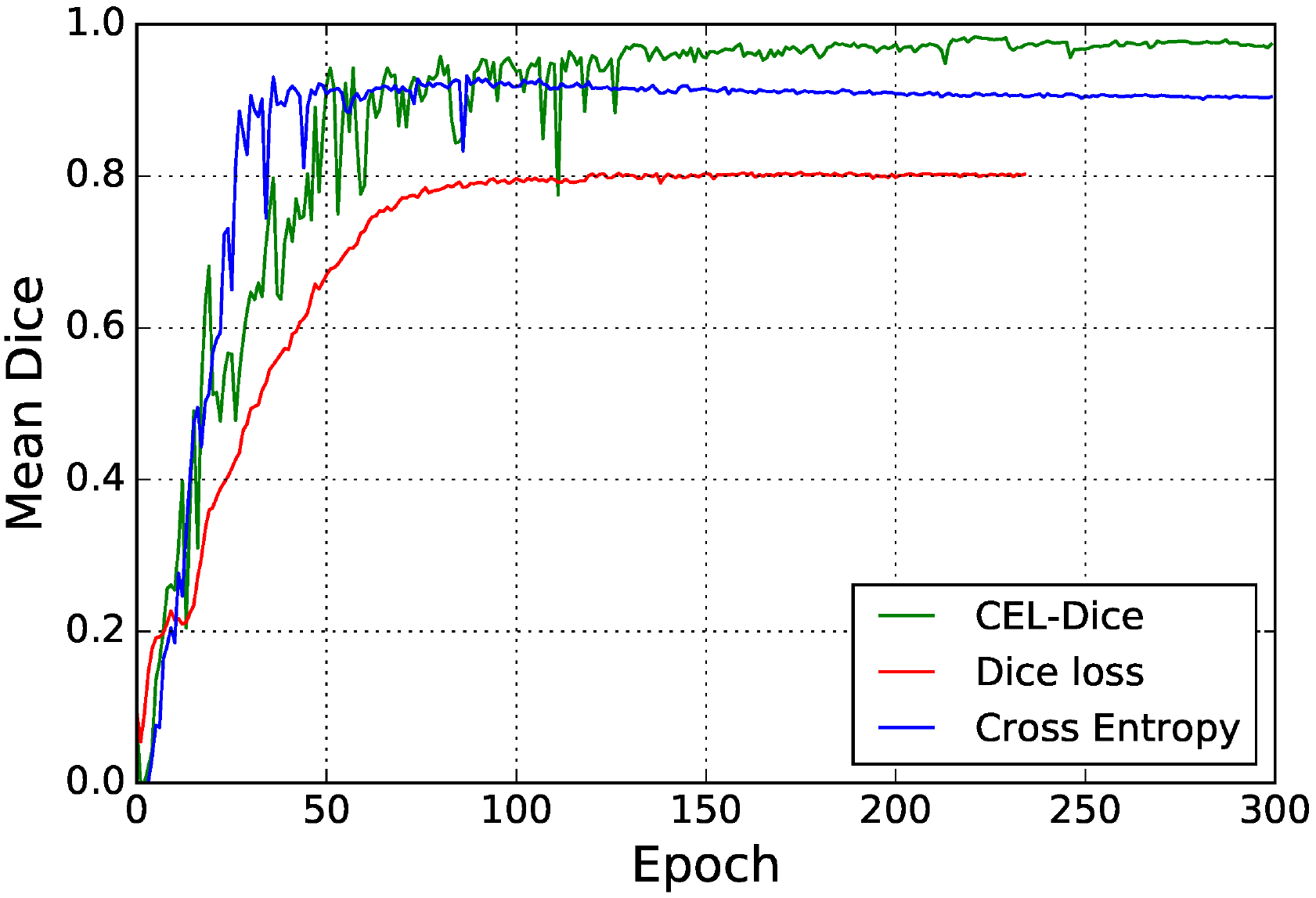}
\end{minipage}
}
\subfigure[The prediction of mean IOU]{
\begin{minipage}{0.46\textwidth}
\centering
\includegraphics[width=\textwidth]{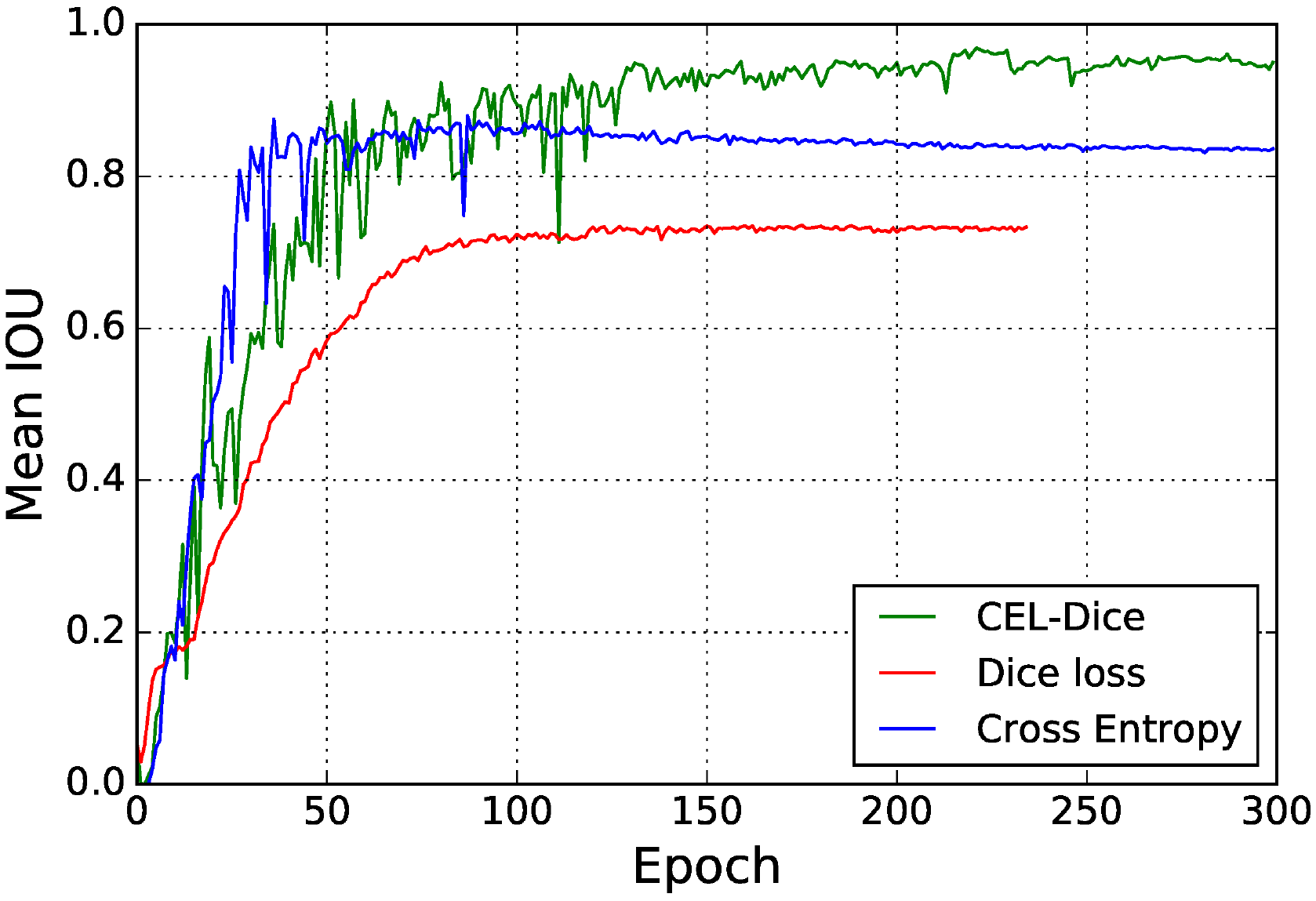}
\end{minipage}
}
\caption{Performance comparison of loss functions. (a) the prediction of mean dice and (b) the prediction of mean IOU.}
\label{dice_compare}
\end{figure}

CEL-Dice is utilized to solve the class balance issue. It combines the stability of cross entropy and the property that Dice loss is not affected by class imbalance. To verify its performance, it is compared with cross entropy and Dice loss. The mean Dice and mean IOU achieved by the network on the test set is illustrated in Fig.~\ref{dice_compare}. They change with the training epoch. It can discover that CEL-Dice can significantly improve segmentation accuracy, which is better than Dice loss and cross entropy.

\section{Conclusion}
A novel network called RAUNet is proposed for semantic segmentation of surgical instruments. The augmented attention module is designed to emphasize key regions. Experimental results show that the augmented attention module can significantly improve segmentation accuracy while adds very few parameters. Also, a hybrid loss called Cross Entropy Log Dice is introduced, contributing to addressing the class imbalance issue. Proved by experiments, RAUNet achieves state-of-the-art performance on Cata7 dataset.

\subsubsection{Acknowledgment}
This research is supported by the National Natural Science Foundation of China (Grants 61533016, U1713220, U1613210), the National Key Research and Development Program of China (Grant 2017YFB1302704) and the Strategic Priority Research Program of CAS (Grant XDBS01040100).

\bibliographystyle{splncs04}
\bibliography{iconip}

\end{document}